\documentclass{article}

\usepackage[final,nonatbib]{nips_2017_arxiv}

\usepackage[utf8]{inputenc} 
\usepackage[T1]{fontenc}    
\usepackage{hyperref}       
\usepackage{url}            
\usepackage{booktabs}       
\usepackage{amsfonts}       
\usepackage{nicefrac}       
\usepackage{microtype}      
\usepackage{bm}
\usepackage{graphicx}

\title{Parametric Gaussian Process Regression for Big Data}

\author{
  Maziar Raissi\thanks{http://www.dam.brown.edu/people/mraissi/} \\
  Division of Applied Mathematics\\
  Brown University\\
  Providence, RI 02912\\
  \texttt{maziar\_raissi@brown.edu} \\
}

\begin{document}

\maketitle

\begin{abstract}
This work introduces the concept of \emph{parametric Gaussian processes} (PGPs), which is built upon the seemingly self-contradictory idea of making Gaussian processes \emph{parametric}. Parametric Gaussian processes, by construction, are designed to operate in ``big data'' regimes where one is interested in \emph{quantifying the uncertainty} associated with noisy data. The proposed methodology circumvents the well-established need for stochastic variational inference, a scalable algorithm for approximating posterior distributions. The effectiveness of the proposed approach is demonstrated using an illustrative example with simulated data and a benchmark dataset in the airline industry with approximately $6$ million records.
\end{abstract}

\section{Introduction} \label{sec:intro}
Gaussian processes (see \cite{Rasmussen06gaussianprocesses, murphy2012machine}) is a non-parametric Bayesian machine learning technique that provides a flexible prior distribution over functions, enjoys analytical tractability, and has a fully probabilistic work-flow that returns robust posterior variance estimates, which quantify uncertainty in a natural way. Moreover, Gaussian processes are among a class of methods known as kernel machines (see \cite{vapnik2013nature, scholkopf2002learning, tipping2001sparse}) and are analogous to regularization approaches (see \cite{tikhonov1963solution, Tikhonov/Arsenin/77, poggio1990networks}). They can also be viewed as a prior on one-layer feed-forward Bayesian neural networks with an infinite number of hidden units \cite{neal2012bayesian}. Non-parametric models such as Gaussian processes need to ``remember'' the full dataset in order to be trained and make predictions. Therefore, the complexity of non-parametric models grows with the size of the dataset. For instance, when applying a Gaussian process to a dataset of size $N$, exact inference has computational complexity $\mathcal{O}(N^3)$ with storage demands of $\mathcal{O}(N^2)$. In recent years, we have been witnessing tremendous amount of efforts (see e.g., \cite{snelson2007local, urtasun2008sparse}) to reduce these complexities. Such efforts generally lead to a computational complexity of $O(N M^2)$ and storage demands of $O(N M)$ where $M$ is a user specified parameter governing the number of ``inducing variables'' (see e.g., \cite{csato2002sparse, seeger2003fast, quinonero2005unifying, titsias2009variational}). However, as is truly pointed out in \cite{hensman2013gaussian} even these reduced storage are prohibitive for ``big data''. In \cite{hensman2013gaussian}, the authors combine the idea of inducing variables with recent advances in variational inference (see e.g., \cite{hensman2012fast, hoffman2013stochastic}) to develop a practical algorithm for fitting Gaussian processes using stochastic variational inference.

In contrast, the current work avoids stochastic variational inference and attempts to present an alternative approach to the one proposed in \cite{hensman2013gaussian}. The seemingly self-contradictory idea is to make Gaussian processes \emph{parametric}. The key feature of parametric models in general, and the current work in particular, is that predictions are conditionally independent of the observed data given the parameters. In other words, the data is distilled into the parameters and any subsequent prediction does not make use of the original dataset. This is very convenient as it enables efficient \emph{mini-batch} training procedures. However, this is not without drawbacks since choosing a model from a particular parametric class constrains its flexibility. Therefore, it is of great importance to devise models that are aware of their imperfections and are capable of properly quantifying the uncertainty in their predictions associated with such limitations.

\section{Methodology}
Let us start by making the prior assumption that
\begin{equation}\label{eq:prior_u}
u(\bm{x}) \sim \mathcal{GP}\left(0, k(\bm{x},\bm{x}';\bm{\theta})\right),
\end{equation}
is a zero mean Gaussian process \cite{Rasmussen06gaussianprocesses} with covariance function $k(\bm{x},\bm{x}';\bm{\theta})$ which depends on the hyper-parameters $\bm{\theta}$. Moreover, let us postulate the existence of some \emph{hypothetical dataset} $\{\bm{Z},\bm{u}\}$ with
\begin{equation}\label{eq:HypotheticalData}
\bm{u} \sim \mathcal{N}(\bm{m},\bm{S}).
\end{equation}
Here, $\bm{Z} = \{\bm{z}^i\}_{i=1}^M$ and $\bm{u} = \{u^i\}_{i=1}^M$. Let us define a \emph{parametric Gaussian process} by the resulting conditional distribution
\begin{equation}\label{eq:PGP}
f(\bm{x}) := u(\bm{x})|\bm{m},\bm{S} \sim \mathcal{GP}\left(\mu(\bm{x};\bm{\theta}, \bm{m}), \Sigma(\bm{x},\bm{x}';\bm{\theta},\bm{S})\right),
\end{equation}
where
\begin{eqnarray}
\mu(\bm{x};\bm{\theta}, \bm{m}) &=& k(\bm{x},\bm{Z};\bm{\theta})k(\bm{Z},\bm{Z};\bm{\theta})^{-1}\bm{m},\label{eq:mu}\label{eq:PGP_mean}\\
\Sigma(\bm{x},\bm{x}';\bm{\theta},\bm{S}) &=& k(\bm{x},\bm{x}';\bm{\theta}) - k(\bm{x},\bm{Z};\bm{\theta})k(\bm{Z},\bm{Z};\bm{\theta})^{-1}k(\bm{Z},\bm{x}';\bm{\theta})\label{eq:k}\label{eq:PGP_variance}\\
&+& k(\bm{x},\bm{Z};\bm{\theta})k(\bm{Z},\bm{Z};\bm{\theta})^{-1} \bm{S} k(\bm{Z},\bm{Z};\bm{\theta})^{-1}k(\bm{Z},\bm{x}';\bm{\theta}).\nonumber
\end{eqnarray}
The parameters $\bm{m}$ and $\bm{S}$ of a parametric Gaussian process (\ref{eq:PGP}) will play a crucial role; The data will be distilled into these parameters and any subsequent predictions will not make use of the original dataset. This is very convenient as it enables an efficient mini-batch training procedure outlined in the following. Taking advantage of the favorable form (\ref{eq:PGP}) of a \emph{parametric Gaussian process}, the mean $\bm{m}$ and covariance matrix $\bm{S}$ of the hypothetical dataset (\ref{eq:HypotheticalData}) can be updated by employing the posterior distribution resulting from conditioning on the observed mini-batch of data $\{\widetilde{\bm{X}},\widetilde{\bm{y}}\}$ of size $\widetilde{N}$; i.e.,
\begin{eqnarray}
\bm{m} &\leftarrow& \mu(\bm{Z};\bm{\theta}, \bm{m}) + \Sigma(\bm{Z},\widetilde{\bm{X}};\bm{\theta}, \bm{S}) \left(\Sigma(\widetilde{\bm{X}},\widetilde{\bm{X}};\bm{\theta}, \bm{S}) + \sigma_\epsilon^2 \bm{I}\right)^{-1}\left[\widetilde{\bm{y}} - \mu(\widetilde{\bm{X}};\bm{\theta},\bm{m})\right],\\
\bm{S} &\leftarrow& \Sigma(\bm{Z},\bm{Z};\bm{\theta}, \bm{S}) - \Sigma(\bm{Z},\widetilde{\bm{X}};\bm{\theta}, \bm{S}) \left(\Sigma(\widetilde{\bm{X}},\widetilde{\bm{X}};\bm{\theta}, \bm{S}) + \sigma_\epsilon^2 \bm{I}\right)^{-1} \Sigma(\widetilde{\bm{X}},\bm{Z};\bm{\theta}, \bm{S}).
\end{eqnarray}
It is worth mentioning that $\mu(\bm{Z};\bm{\theta}, \bm{m}) = \bm{m}$ and $\Sigma(\bm{Z},\bm{Z};\bm{\theta}, \bm{S}) = \bm{S}$. The information corresponding to the mini-batch $\{\widetilde{\bm{X}},\widetilde{\bm{y}}\}$ is now distilled in the parameters $\bm{m}$ and $\bm{S}$. The hyper-parameters $\bm{\theta}$ and noise variance parameter $\sigma_\epsilon^2$ can be updated by taking a step proportional to the gradient of the \emph{negative log marginal likelihood}
\begin{eqnarray}
\mathcal{NLML}(\bm{\theta}, \sigma_\epsilon^2) := \frac{1}{2} \bm{m}^T k(\bm{Z}, \bm{Z}; \bm{\theta})^{-1} \bm{m} + \frac{1}{2} \log |k(\bm{Z}, \bm{Z}; \bm{\theta})| + \frac{1}{2} M \log (2 \pi).
\end{eqnarray}
The training procedure is initialized by setting $\bm{m}_0 = \bm{0}$ and $\bm{S}_0 = k(\bm{Z},\bm{Z};\bm{\theta}_0)$ where $\bm{\theta}_0$ is some initial set of hyper-parameters. Having trained the hyper-parameters and parameters of the model, one can use equation (\ref{eq:mu}) to predict the mean $\mu(\bm{x}^*;\bm{\theta},\bm{m})$ of the solution at a new test point $\bm{x}^*$. Moreover, the predicted variance is given by $\Sigma(\bm{x}^*,\bm{x}^*;\bm{\theta},\bm{S})$, where $\Sigma$ is obtained from equation (\ref{eq:k}).

\section{Experiments}
Parametric Gaussian process regression is entirely agnostic to the size of the dataset and can effectively handle datasets with millions or billions of records. The effectiveness of the proposed methodology will be demonstrated using an illustrative example with simulated data and a benchmark dataset in the literature on Gaussian processes and big data.

\begin{figure}
\centering
\includegraphics[width=0.85\linewidth]{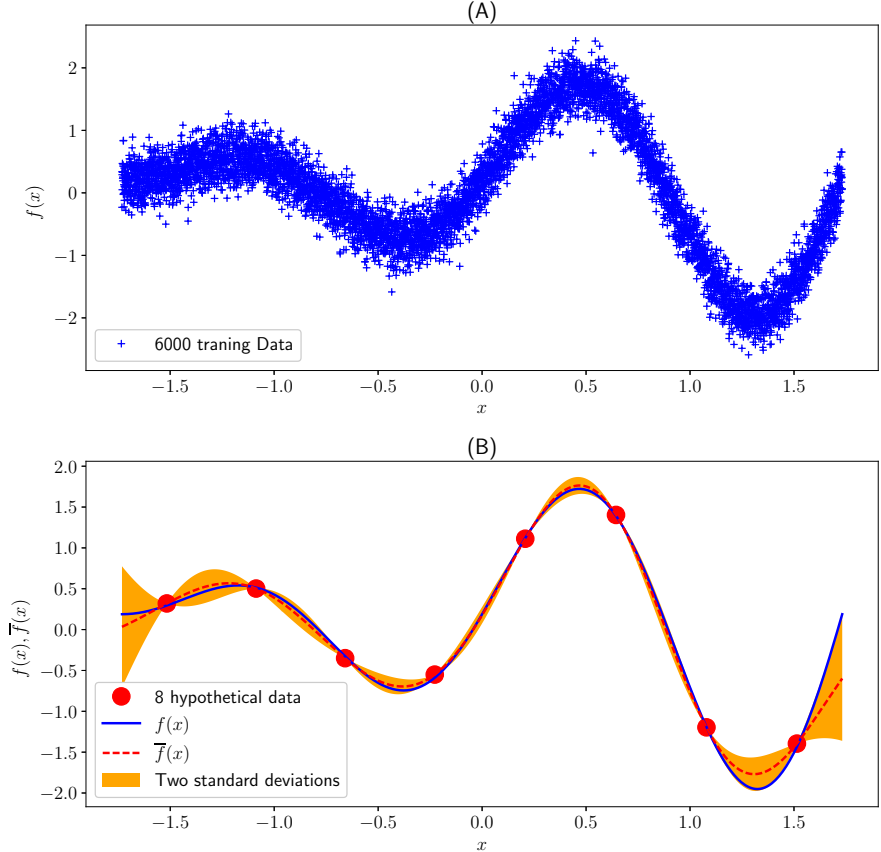}
\caption{\emph{Illustrative example:} (A) Plotted are 6000 training data generated by random perturbations of the one dimensional function $f(x) = x \sin(4 \pi x)$. (B) Depicted is the resulting prediction of the model. The blue solid line represents the true data generating function $f(x)$, while the dashed red line depicts the predicted mean $\overline{f}(x)$. The shaded orange region illustrates the two standard deviations band around the mean. The red circles depict the resulting mean values $\bm{m}$ for the 8 hypothetical data points $\{\bm{Z},\bm{u}\}$ after a pass through the entire dataset while mini-batches of size one are employed per each iteration of the training algorithm. It is remarkable how the training procedure places the mean $\bm{m}$ of the hypothetical dataset on the underlying function $f(x)$. {\it (Code: \url{http://bit.ly/2qwR5eW})}}\label{fig:OneDimensional}
\end{figure}

\subsection{Illustrative example}
To demonstrate the proposed framework, let us begin with a simple dataset generated by random perturbations of a one dimensional function given explicitly by $f(x) = x \sin(4 \pi x)$. The $6000$ training data are depicted in panel (A) of figure \ref{fig:OneDimensional}. The Gaussian process prior (\ref{eq:prior_u}) used for this example is assumed to have a squared exponential \cite{Rasmussen06gaussianprocesses} covariance function, i.e.,
\begin{eqnarray*}
k(x,x';\bm{\theta}) = \gamma^2 \exp\left(-\frac12 w^2(x - x')^2\right),
\end{eqnarray*}
where $\gamma^2$ is a variance parameter and $\bm{\theta} = \left(\gamma, w\right)$ are the hyper-parameters. The model employs a \emph{hypothetical data-set} (see equation (\ref{eq:HypotheticalData})) of size $M=8$. The locations $\bm{Z}$ of the hypothetical dataset are obtained by employing the k-means clustering algorithm. The training procedures is carried out using the Adam stochastic optimizer \cite{kingma2014adam} with default settings and mini-batches of size one. After one pass through the entire training data, it is remarkable how the parameters $\bm{m}$ and $\bm{S}$ of the hypothetical dataset enable us to summarize the actual training data. The red circles in figure \ref{fig:OneDimensional} denote the pairs $\{\bm{Z}, \bm{m}\}$ of the hypothetical data. The resulting prediction of the model is plotted in figure \ref{fig:OneDimensional}.

\subsection{Airline delays}
\begin{figure}
\centering
\includegraphics[width=\textwidth]{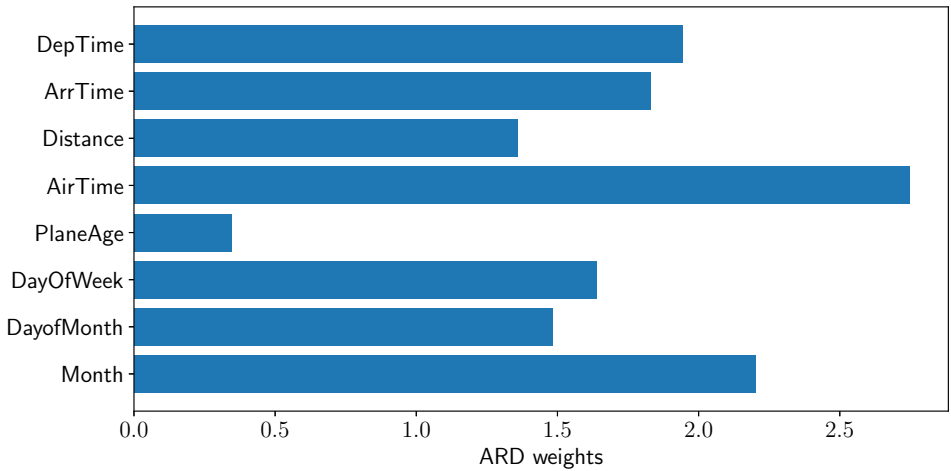}
\caption{\emph{Airline delays example:} Automatic relevance determination parameters for the features used for predicting flight delays. {\it (Code: \url{http://bit.ly/2qwR5eW})}}\label{fig:Airline_ARD}
\end{figure}

The US flight delay prediction example, originally proposed in \cite{hensman2013gaussian}, has reached a status of a standard benchmark dataset (see e.g., \cite{hensman2016variational, gal2014distributed, DBLP:conf/icml/DeisenrothN15, samo2016string, adam2016scalable}) in Gaussian process regression, partly because of the massive size of the dataset with nearly $5.93$ million records and partly because of its large-scale non-stationary nature. The dataset\footnote{\href{http://stat-computing.org/dataexpo/2009/}{http://stat-computing.org/dataexpo/2009/}} consists of flight arrival and departure times for every commercial flight in the USA for the year 2008. Each record is complemented with details on the flight and the aircraft. The aim is to predict the delay in minutes of the aircraft at landing, $y$. The eight covariates $\bm{x}$ are the same as \cite{hensman2013gaussian}, namely the age of the aircraft (number of years since deployment), route distance, airtime, departure time, arrival time, day of the week, day of the month, and month. Two third of the entire data set, which totals $3.95$ million records, is used for training  and one third for testing. The output data are normalized by subtracting the training sample mean from the outputs and dividing the results by the sample standard deviation. The input data are normalized to the interval $[0, 1]$. The Gaussian process prior (\ref{eq:prior_u}) used for this example is assumed to have a squared exponential \cite{Rasmussen06gaussianprocesses} covariance function, i.e.,
\begin{eqnarray*}
k(\bm{x},\bm{x}';\bm{\theta}) = \gamma^2 \exp\left(-\frac12\sum_{d=1}^8 w_{d}^2(x_d - x'_d)^2\right),
\end{eqnarray*}
where $\gamma^2$ is a variance parameter, $\bm{x}$ is the vector of covariates, and $\bm{\theta} = \left(\gamma, w_1, \ldots, w_8\right)$ are the hyper-parameters. Moreover, anisotropy across input dimensions is handled by Automatic Relevance Determination (ARD) weights $w_{d}$. From a theoretical point of view, each kernel gives rise to a Reproducing Kernel Hilbert Space \cite{aronszajn1950theory, saitoh1988theory, berlinet2011reproducing} that defines a class of functions that can be represented by this kernel. In particular, the squared exponential covariance function chosen above implies smooth approximations. More complex function classes can be accommodated by appropriately choosing kernels. The model employs a \emph{hypothetical dataset} (see equation (\ref{eq:HypotheticalData})) of size $M=500$. The locations $\bm{Z}$ of the hypothetical dataset are obtained by employing the k-means clustering algorithm. The training procedures is carried out using the Adam stochastic optimizer \cite{kingma2014adam} with default settings and mini-batches of size $1000$. After $10000$ iterations of the training procedure, the predictive mean squared error (MSE) on the normalized test data is given by $0.832810$. This value for the MSE is within the range reported in the literature (see e.g., table 2 in \cite{hensman2016variational}). The MSE over the normalized data can be interpreted as a fraction of the sample variance of airline arrival delays. Thus a MSE of $1.00$ is as good as using the training mean as predictor. In order to further reduce the MSE one could increase the size $M$ of the \emph{hypothetical} data-set, increase the batch-size, and/or choose a more accommodative covariance function. Moreover, to get a better idea of the relevance of the different features available in this dataset, figure \ref{fig:Airline_ARD} plots the automatic relevance determination parameters $w_d$. The most relevant variable turns out to be the airtime that needs to be covered. The month and time of departure of the flight are also two important features in predicting flight delays.

\section{Related works}
Despite some subtle differences, it is generally safe to recognize the input-output pairs $\{\bm{Z},\bm{u}\}$ (see equation (\ref{eq:HypotheticalData})) as the so called ``inducing points'', a frequently used term in the literature on sparse approximations to Gaussian process priors (see e.g., \cite{quinonero2005unifying} for a compressive review). However, it is not advisable to interpret $\bm{m}$ and $\bm{S}$ (see equation (\ref{eq:HypotheticalData})) as variational parameters \cite{hensman2013gaussian} since no (stochastic) variational inference is carried out in the current work. Furthermore, to highlight the subtle differences between ``inducing points'' and what this work calls \emph{hypothetical dataset} (\ref{eq:HypotheticalData}), it is worth observing that in the literature on sparse approximations to Gaussian processes it turns out that $\bm{m} = \bm{0}$ and $\bm{S} = k(\bm{Z},\bm{Z};\bm{\theta})$. Under these assumptions and using equations (\ref{eq:PGP_mean}) and (\ref{eq:PGP_variance}), one obtains $\mu(\bm{x};\bm{\theta}, \bm{m}) = 0$ and $\Sigma(\bm{x},\bm{x}';\bm{\theta},\bm{S}) = k(\bm{x},\bm{x}';\bm{\theta})$. In other words, in the sparse Gaussian processes framework, $f(\bm{x})$ and $u(\bm{x})$ are essentially identical; i.e., $f(\bm{x}) = u(\bm{x}) \sim \mathcal{GP}\left(0, k(\bm{x},\bm{x}';\bm{\theta})\right)$. In contrast, this work treats $\bm{m}$ and $\bm{S}$ as parameters of the model responsible for encoding the history of observed data. In this regard, the current work is similar to \cite{hensman2013gaussian}. However, unlike \cite{hensman2013gaussian}, the parameter $\bm{m}$ and $\bm{S}$ are not variational parameters of some variational distribution.

\section{Concluding remarks}
Modern datasets are rapidly growing in size and complexity, and there is a pressing need to develop new statistical methods and machine learning techniques to harness this wealth of data. This work presented a novel regression framework for encoding massive amount of data into a small number of \emph{hypothetical} data points. While being effective, the resulting model is conceptually very simple, is based on the idea of making Gaussian processes \emph{parametric}, and it takes at most $8$ mathematical formulas to explain every single detail of the algorithm. This simplicity is extremely important specially when it comes to deploying machine learning algorithms on big data flow engines (see e.g., \cite{meng2016mllib}) such as MapReduce \cite{dean2008mapreduce} and Apache Spark \cite{zaharia2012resilient}. Moreover, Gaussian processes are a powerful tool for probabilistic inference over functions. They offer desirable properties such as uncertainty estimates, automatic discovery of important dimensions, robustness to over-fitting, and principled ways of tuning hyper-parameters. Thus, scaling Gaussian processes to big datasets and deploying it on big data flow engines is, and will remain, an active area of research.

\subsubsection*{Acknowledgments}
This works received support by the DARPA EQUiPS grant N66001-15-2-4055, and the AFOSR grant FA9550-17-1-0013.

\small
\bibliographystyle{unsrt}
\bibliography{sample.bib}

\end{document}